\documentclass[letterpaper, 10 pt, conference]{ieeeconf}

\usepackage{times}
\usepackage{graphicx}
\usepackage{amsmath,amssymb,amsopn,amstext,amsfonts}
\usepackage{cancel}
\usepackage[space]{cite}
\usepackage{balance}
\usepackage{color}
\usepackage{mathtools}
\usepackage{algpseudocode}
\usepackage{algorithm}
\usepackage{bm}

\usepackage{diagbox}
\usepackage{float}
\usepackage{pifont}
\usepackage{multirow}
\usepackage{url}
\usepackage{verbatim}
\usepackage{booktabs}
\usepackage{capt-of}
\usepackage{threeparttable}
\usepackage{makecell}
\usepackage{soul}
\usepackage{xcolor}
\usepackage[normalem]{ulem}
\usepackage{etoolbox}
\usepackage{mathrsfs}
\usepackage{hyperref}
\usepackage{stfloats}
\hypersetup{
	colorlinks=true,
	linkcolor=black,
	anchorcolor=black,
	citecolor=black,
    urlcolor=black}
\newcommand{\projectwebsite}{\url{https://junxiaolin.github.io/poise-website/}}

\graphicspath{{./figures/}}
\DeclareGraphicsExtensions{.png,.jpg,.eps,.pdf,.emf}
\IEEEoverridecommandlockouts
\hypersetup{
    pdftitle={Learning In-Hand Object Reaching to General 6D Poses},
    pdfauthor={Junxiao Lin, Tianyue Wu, Jie Yin, Jia Pan, Kaifeng Zhang, Weiming Zhi}
}

\begin{document}
	\title{\LARGE \bf Learning In-Hand Object Reaching to General 6D Poses}
    \author{Junxiao Lin$^{2,1,3}$, Tianyue Wu$^{2,3}$, Jie Yin$^{2}$, Jia Pan$^3$, Kaifeng Zhang$^2$, Weiming Zhi$^{1,*}$
	\thanks{$^1$ School of Computer Science, The University of Sydney, Australia.
    $^2$ Sharpa.
	$^3$ School of Computing and Data Science, The University of Hong Kong, HKSAR.
    $^*$ Corresponding author.}}

    \IEEEaftertitletext{%
        \begin{minipage}{\textwidth}
            \centering
            \includegraphics[width=\textwidth]{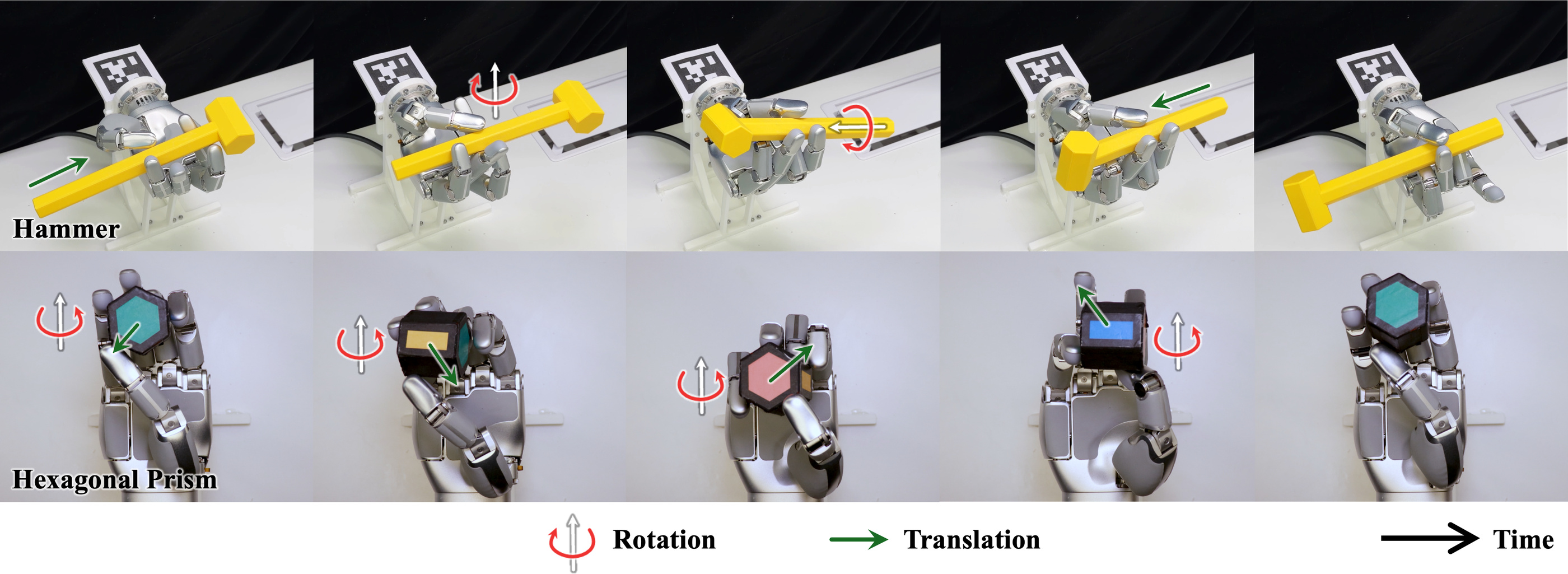}
            \captionof{figure}{Real-world in-hand 6D pose reaching. The Hammer
            (top) and Hexagonal Prism (bottom) are translated and rotated within
            the hand toward commanded poses. Green and red arrows indicate
            translational and rotational motions, respectively.}
            \label{fig:real_target_tracking}
            \vspace{0.5\baselineskip}
        \end{minipage}}

\maketitle
	\thispagestyle{empty}
	\pagestyle{empty}

	\begin{abstract}
		\label{sec:abstract}
        In-hand manipulation allows multi-fingered dexterous hands to
        reconfigure grasped objects without releasing and regrasping them. This
        improves manipulation efficiency by reducing repeated grasp acquisition
        and large arm motions. However, most learning-based methods focus on
        reorientation, continuous rotation, or translation, whereas
        many tasks require joint control of object position and orientation. We
        formulate this capability as in-hand 6D object pose reaching: starting
        from an existing grasp, coordinated finger motions move the object to a
        palm-relative target pose. We present
        \textbf{POISE} (\textbf{P}alm-relative \textbf{O}bject reaching
        \textbf{I}n $\mathbf{SE}(3)$), a sim-to-real reinforcement learning
        framework for this task. POISE combines diverse stable-grasp
        initialization, goal- and geometry-conditioned control, an adaptive 6D
        goal curriculum, and a compact reward scheme for pose reaching and grasp
        preservation. In simulation, diverse initialization raises held-out-grasp success
        from 40.1\% to 51.5\% and post-drop recovery from 33.8\% to 72.9\%; the
        curriculum raises full-range success from 6.2\% to 59.5\%.
        On hardware, the grasp-maintenance reward improves three-target
        sequence success from 20\% to 80\%. In
        real-world experiments, POISE reaches successive 6D targets without
        manual reset across multiple object geometries and wrist orientations, and
        recovers from external disturbances.
        To support further research in dexterous manipulation, we will
        release our code at \projectwebsite.

	\end{abstract}

	\IEEEpeerreviewmaketitle

	\section{Introduction}
	\label{sec:Introduction}
    Many manipulation tasks begin, rather than end, once an object has been grasped. While parallel-jaw grippers are effective for establishing stable grasps, they have limited ability to reconfigure an object after grasp acquisition. High-degree-of-freedom dexterous hands, in contrast, can continue to manipulate a grasped object through coordinated multi-finger motions. This in-hand dexterity is crucial when a robot needs to shift the grasping region on a tool, expose a desired contact surface, or align a functional axis with the environment.

    We formulate this capability as in-hand 6D object pose reaching: given a grasped object and a random target pose defined relative to the hand in the special Euclidean group $\mathrm{SE}(3)$, a single multi-fingered dexterous hand must bring the object to the target using only in-hand manipulation. Unlike in-hand reorientation or translation-only tasks~\cite{andrychowicz2020learning,qi2025translation}, this formulation jointly constrains the object's relative 3D translation and 3D rotation. This coupled objective significantly constrains the feasible motion space and often requires contact switching or finger gaiting while maintaining a stable grasp.
    Successive reaching further requires each attained object pose to be
    supported by a grasp that permits continued manipulation.

    Prior work has addressed aspects of this problem from two main directions. Model-based methods can achieve accurate local in-hand pose control, but they often rely on predefined contact modes, fixed contact sequences, or known and simplified object geometry~\cite{morgan2022complex}. As a result, they do not readily scale to random $\mathrm{SE}(3)$ reaching, where the hand must actively switch contacts and adapt to diverse object shapes. Reinforcement learning (RL), on the other hand, has become a powerful tool for dexterous in-hand manipulation because it can discover contact-rich behaviors without requiring high-quality demonstrations, which are difficult to acquire for such tasks. However, most learning-based in-hand manipulation work focuses on reorientation, continuous rotation, or constrained translation~\cite{andrychowicz2020learning,qi2023hora,chen2023visual,qi2025translation}. Prior full-pose in-hand reaching methods remain confined to object-specific simulation~\cite{plappert2018multigoal,charlesworth2021solving}.

    To bridge these gaps, we develop \textbf{POISE} (\textbf{P}alm-relative
    \textbf{O}bject reaching \textbf{I}n $\mathbf{SE}(3)$), a sim-to-real
    RL framework based on Proximal Policy Optimization
    (PPO)~\cite{schulman2017ppo}. POISE comprises five components: diverse
    physics-validated stable-grasp resets for broad state
    coverage; a goal- and geometry-conditioned policy for shape-dependent
    control; an adaptive 6D goal curriculum for learning large pose changes; a
    compact reward scheme that combines pose reaching, goal completion, grasp maintenance,
    and regularization; and domain randomization for robust sim-to-real transfer.
    Experiments demonstrate generalization across initial grasps, recovery
    after the original grasp is lost,
    and continued 6D reaching across multiple object geometries and wrist
    orientations.

    The main contributions of this letter are as follows:
    \begin{enumerate}
    \item A diverse, physics-validated stable-grasp initialization that
    improves generalization to unseen grasps and recovery after the original
    grasp is lost.

    \item An RL recipe for in-hand 6D pose reaching that combines
    geometry-conditioned control and adaptive goal curricula with a
    grasp-maintenance reward based on balanced wrench coverage.

    \item Real-world validation of successive 6D reaching across object
    geometries and wrist orientations, including disturbance recovery
    and a quantitative evaluation of the grasp-maintenance reward.
    \end{enumerate}

    \section{Related Work}
    \label{sec:related_work}

    Dexterous in-hand manipulation has been extensively studied, from early
    analyses of rolling and finger gaiting~\cite{han1998rolling} to modern
    model-based and learning-based approaches.
    Model-based methods compute manipulation motions from explicit models of
    hand kinematics, fingertip contacts, and object motion. Motion-cone
    planning derives feasible object motions from contact
    constraints~\cite{dafle2018motioncones}, while trajectory optimization
    and multi-modal planning coordinate in-grasp motion, finger gaiting, and
    compliant contact transitions~\cite{sundaralingam2019relaxed,
    morgan2022complex}. Model predictive control instead plans over locally
    identified or contact-implicit dynamics~\cite{chanrungmaneekul2023nonparametric,
    jiang2024contact,suh2026contacttrust}. These approaches can enforce physical and kinematic
    constraints directly, but often depend on accurate models, predefined
    contact modes, or time-consuming online optimization. Extending them to a broad
    range of 6D goals and object geometries remains challenging.

    Learning-based methods instead discover contact-rich multi-finger
    coordination through trial-and-error
    interaction~\cite{khandate2023sampling,yin2025dexteritygen,zhang2026unicross,
    pei2026assembling}. RL has enabled real-world
    reorientation, continuous rotation, and constrained translation~\cite{akkaya2019rubiks,andrychowicz2020learning,
    chen2022system,handa2023dextreme,qi2023hora,qi2023rotateit,yin2023rotating,
    yang2024anyrotate,liu2026dexndm,qi2025translation,bhardwaj2026viserdex,
    yin2026wmcraftnet}. Recent
    work further improves generalization across object geometry through
    visual or explicit shape conditioning~\cite{huang2021geometry,
    chen2023visual,pitz2024shapeconditioned}. Nevertheless, most of these
    methods control orientation or a constrained translational motion rather
    than a general 6D target pose.

    Closest to our setting, Plappert et al.~\cite{plappert2018multigoal}
    introduced goal-pose-conditioned RL simulation environments for in-hand
    manipulation of block, egg, and pen objects. Initial solutions to these
    challenging tasks achieved only low success rates; Charlesworth and
    Montana~\cite{charlesworth2021solving} later improved performance using
    offline trajectory optimization, but still only in simulation. More recent systems learn 6D object pose reaching
    through joint arm--hand control toward world-frame
    goals~\cite{kedia2026simtoolreal,lum2026play2perfect,kuang2026dex4d}.
    Other controllers track human or synthetic hand--object reference
    motions~\cite{liu2025dextrack,li2025maniptrans,wang2025handobjecttracking,li2026teledexter}.
    In contrast, POISE reaches palm-relative $\mathrm{SE}(3)$ object goals
    through finger motions alone, without hand-motion references.

    \section{Problem Formulation}
    \label{sec:problem_formulation}

    We consider a fixed-wrist hand moving an already-grasped object to
    commanded poses using only finger motions. Let $\mathcal H$ and
    $\mathcal O$ denote the palm- and object-fixed frames. The current and
    target palm-relative poses are
    ${}^{\mathcal H}\mathbf T_{\mathcal O,t}
    =({}^{\mathcal H}\mathbf p_t,{}^{\mathcal H}\mathbf R_t)$ and
    ${}^{\mathcal H}\mathbf T_{\mathcal O,g}
    =({}^{\mathcal H}\mathbf p_g,{}^{\mathcal H}\mathbf R_g)\in\mathrm{SE}(3)$.
    Their position and orientation errors are
    \begin{equation}
        \begin{aligned}
        e_p &= \bigl\|{}^{\mathcal H}\mathbf p_t
        -{}^{\mathcal H}\mathbf p_g\bigr\|_2,\\
        e_R &= d_{\mathrm{SO}(3)}\!\left(
        {}^{\mathcal H}\mathbf R_t,{}^{\mathcal H}\mathbf R_g\right)\\
        &= \frac{1}{\sqrt{2}}
        \left\|\operatorname{Log}\!\left(
        {}^{\mathcal H}\mathbf R_g
        ({}^{\mathcal H}\mathbf R_t)^\top
        \right)\right\|_{\mathrm F}.
        \end{aligned}
        \label{eq:pose_errors}
    \end{equation}

    Here, $\operatorname{Log}(\cdot)$ denotes the matrix logarithm and
    $\|\cdot\|_{\mathrm F}$ denotes the Frobenius norm; thus, $e_R\in[0,\pi]$ is the
    geodesic rotation angle between the two orientations.

    We formulate the task as a goal-conditioned partially observable Markov
    decision process (POMDP). An MLP actor receives three-frame histories of
    the current and commanded finger-joint positions, estimated palm-frame
    object pose and velocity, the gravity direction expressed in the palm frame,
    a confidence score for the visual pose estimate, the target pose, the current-to-target translation
    and rotation, and an object-geometry descriptor $\mathbf z_{\mathcal O}$
    (Sec.~\ref{sec:geometry_policy}).
    The critic additionally receives noise-free simulator
    states, fingertip states, applied joint torques, and contact forces.
    The normalized action incrementally updates the commanded finger joints,
    \begin{equation}
        \mathbf q^{\mathrm{cmd}}_{t+1}=\operatorname{clip}\!\left(
        \mathbf q^{\mathrm{cmd}}_t+s_a\mathbf a_t,
        \mathbf q_{\min},\mathbf q_{\max}\right).
        \label{eq:action}
    \end{equation}

    A target is reached when both errors remain within their tolerances for a
    prescribed number of control steps. A new target is then sampled without
    resetting the hand--object state during training, requiring successive 6D
    reaches within one continuous rollout.

    \section{Method}
    \label{sec:method}

\begin{figure}[t]
        \centering
        \includegraphics[width=\columnwidth]{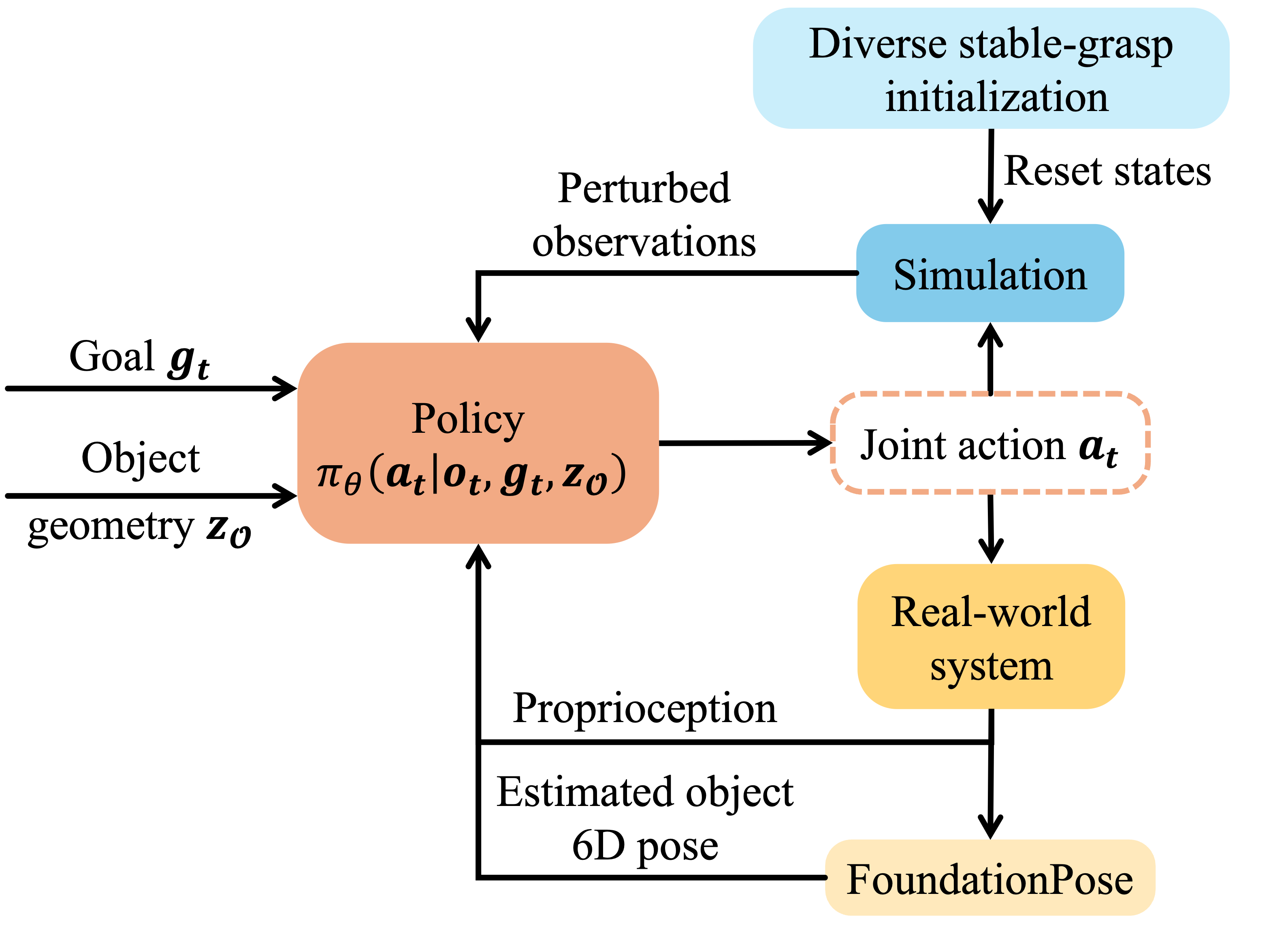}
        \caption{Overview of POISE. Diverse stable grasps initialize simulation
        training, while FoundationPose supplies object-pose feedback during
        real-world deployment.}
        \label{fig:pipeline}
    \end{figure}

    In-hand 6D pose reaching requires the policy to handle varied grasps and object
    geometries, explore a broad range of 6D poses, and preserve a
    stable grasp after reaching. POISE addresses these challenges through
    diverse stable-grasp initialization, geometry-conditioned control, an
    adaptive 6D goal curriculum, a grasp-preserving reward, and domain randomization, as summarized in Fig.~\ref{fig:pipeline}.

    \subsection{Diverse Stable-Grasp Initialization}
    \label{sec:grasp_initialization}

    Because the task starts after grasp acquisition, each episode must begin
    from a stable hand--object state. With only one reset grasp, rollout
    exploration is largely restricted to states that the current policy can
    reach from that configuration, leaving alternative object--palm poses and
    contact arrangements underrepresented. We therefore initialize training
    from a diverse set of stable grasps.

    Following the optimization-based grasp synthesis paradigm~\cite{wang2023dexgraspnet},
    we generate these grasps using a procedure adapted from~\cite{yin2025lightning}.
    Object poses are first sampled over a feasible
    in-hand workspace. For each pose, we optimize stable contacts and solve for
    the corresponding joint configuration $\mathbf q$; the optimized contact
    forces are then converted into an executable position-control command
    $\mathbf q^{\mathrm{cmd}}$. Randomized contact search produces multiple
    postures for each object pose. We then perturb the object pose,
    $\mathbf q$, and $\mathbf q^{\mathrm{cmd}}$, and simulate each candidate
    for 2~s under gravity and position control. Only candidates that retain the
    object are stored in the reset cache $\mathcal G$.

    Sampling reset states from $\mathcal G$ makes grasp diversity an explicit part of
    the training distribution, rather than leaving it to be discovered through
    rollout exploration. We evaluate its effect on held-out grasps and recovery
    in Sec.~\ref{sec:grasp_initialization_ablation}.

    \subsection{Geometry-Conditioned Policy}
    \label{sec:geometry_policy}

    Object shape determines which surfaces and edges can support contact and
    how finger motion is transferred to the object. Thus, identical current and
    target poses can require different finger coordination for different
    geometries. We expose this information through an explicit shape
    descriptor.

    We use the basis point set (BPS) representation~\cite{prokudin2019efficient,
    pitz2024shapeconditioned}. A fixed set of query
    points is shared by all objects in their canonical frames, and each entry
    records the normalized distance from one query point to the nearest object
    surface. The resulting ordered vector $\mathbf z_{\mathcal O}$ has the same
    dimension and spatial structure across objects.

    The actor receives $\mathbf z_{\mathcal O}$ together with its observation
    and goal. During multi-object training, one actor is optimized jointly
    across objects without a categorical identity input; it must therefore use
    geometry to adapt its finger coordination. The BPS dimension is given in
    Sec.~\ref{sec:experimental_setup}.

    \subsection{Adaptive 6D Goal Curriculum}
    \label{sec:curriculum}

    Sampling the full range of 6D goals from the outset makes early exploration
    difficult. We therefore use separate curricula that expand the maximum
    rotation and translation magnitudes from $5^\circ$ and 10~mm to $180^\circ$
    and 30~mm, respectively. Only the magnitudes are scheduled; axes and
    directions remain randomized, with the same limits applied to all targets
    in a rollout. Targets outside the palm-frame workspace or intersecting the
    fixed hand base are resampled.

    At each stage, magnitudes are sampled near the frontier, from the exposed
    range, or at the current limit in a $0.6/0.3/0.1$ ratio. When frontier
    success reaches 0.4, the rotation or translation limit increases by
    $10^\circ$ or 10~mm. Each object maintains independent curriculum progress.

    \subsection{Grasp-Preserving Reward Design}
    \label{sec:reward}

    Reaching the target pose does not guarantee a useful terminal grasp: the
    object may be loosely supported or held by too few contacts to continue
    manipulation. Our reward therefore combines pose reaching, verified goal
    completion, grasp maintenance, and actuation regularization.

    \noindent\textbf{Pose reaching.}
    Using the errors in Eq.~\eqref{eq:pose_errors}, the dense pose reward is
    \begin{gather}
        r_t^{R} = \exp\!\Bigl(-\frac{e_{R,t}}{30^\circ}\Bigr), \\
        r_t^{p} = \alpha(e_{R,t})\,
        \exp\!\Bigl(-\frac{e_{p,t}}{\sigma(e_{R,t})}\Bigr), \\
        r_t^{\mathrm{pose}} = \tfrac{1}{6}r_t^{R}
        +\tfrac{1}{4}r_t^{p},
        \label{eq:pose_reward}
    \end{gather}
    where $\alpha$ and $\sigma$ vary with $e_R$ as
    \begin{equation}
        (\alpha,\sigma)=
        \begin{cases}
            (0.15,\,30~\mathrm{mm}), & e_R>45^\circ,\\
            (0.40,\,20~\mathrm{mm}), & 25^\circ<e_R\leq45^\circ,\\
            (1.00,\,15~\mathrm{mm}), & e_R\leq25^\circ.
        \end{cases}
        \label{eq:position_shaping_schedule}
    \end{equation}
    When the orientation error is large, the position term is broad and
    downweighted, allowing translation needed for contact rearrangement. As the
    object becomes rotationally aligned, the reward increasingly emphasizes
    accurate positioning.

    \noindent\textbf{Goal completion.}
    Let $s_t\in\{0,1\}$ indicate that the object is retained and lies within
    both pose tolerances. A goal is verified only after this condition holds for
    $N=20$ consecutive steps; $c_t\in\{0,1\}$ marks the first step on which
    verification occurs. We use
    \begin{equation}
        r_t^{\mathrm{goal}}=0.5s_t+45c_t.
        \label{eq:goal_reward}
    \end{equation}
    The per-step term encourages the policy to remain inside the target region,
    while the one-time bonus rewards verified completion.

    \noindent\textbf{Grasp maintenance.}
    We approximate each active contact by four friction-cone wrench rays, with
    moments normalized by object size. For each force and torque axis,
    $m_{t,k}\in[0,1]$ is the smaller of the maximum ray projections in its
    positive and negative directions. We aggregate the six margins as
    \begin{equation}
        Q_t=\left[\frac{1}{6}\sum_{k=1}^{6}
        (m_{t,k}+\varepsilon)^{-8}\right]^{-1/8}-\varepsilon.
        \label{eq:grasp_wrench_quality}
    \end{equation}
    This generalized mean emphasizes the weakest of the six bidirectional
    projection margins, encouraging balanced wrench coverage along the three
    force and three torque axes. Contact
    contributions saturate with force, preventing the policy from increasing
    the score merely by squeezing harder.

    Since attainable quality depends on the object and initial grasp, the reward
    preserves quality relative to the episode's stable reset rather than using
    one absolute threshold. Let $Q^\star$ be the mean quality over the first
    four control steps, clipped to $[0.08,0.35]$. We define
    \begin{equation}
        r_t^{\mathrm{grasp}}
        =1-\operatorname{clip}\!\left(
        \frac{(Q^\star-Q_t)_+}{Q^\star},0,1\right)^2.
        \label{eq:grasp_maintenance_reward}
    \end{equation}
    The reward remains maximal while the initial wrench coverage is retained
    and decreases smoothly as the grasp becomes less secure.

    \noindent\textbf{Drop and actuation regularization.}
    Let $d_t\in\{0,1\}$ indicate a drop. To discourage object loss and
    unnecessarily aggressive control, the complete reward penalizes drops,
    joint torque, and instantaneous mechanical power:
    \begin{equation}
        \begin{aligned}
        r_t &= r_t^{\mathrm{pose}}+r_t^{\mathrm{goal}}
        +0.05r_t^{\mathrm{grasp}}\\
        &\quad -80d_t-0.10\|\boldsymbol\tau_t\|_2^2
        -0.15\bigl(\boldsymbol\tau_t^\top\dot{\mathbf q}_t\bigr)^2.
        \end{aligned}
        \label{eq:total_reward}
    \end{equation}

    \subsection{Domain Randomization}
    \label{sec:training_transfer}

    Real deployment introduces both dynamics mismatch and imperfect visual pose
    feedback. We randomize object properties, friction, controller gains, and
    external disturbances to improve robustness to the former. For the latter,
    the actor receives joint and object-pose noise, pose delay, and hold-last
    dropouts; object velocities are estimated from the corrupted pose history.
    Table~\ref{tab:sim2real_randomization} summarizes the ranges.

    \begin{table}[H]
        \centering
        \caption{Sim-to-real randomization used during training. $\mathcal U$
        and $\mathrm{LogU}$ denote uniform and log-uniform sampling.}
        \label{tab:sim2real_randomization}
        \setlength{\tabcolsep}{3pt}
        \renewcommand{\arraystretch}{1.03}
        \footnotesize
        \begin{tabular}{@{}>{\raggedright\arraybackslash}p{0.35\columnwidth}>{\raggedright\arraybackslash}p{0.59\columnwidth}@{}}
            \toprule
            Quantity & Range or distribution \\
            \midrule
            Object scale & $\mathcal U[0.975,1.025]$ \\
            Object mass & $10$--$100$~g \\
            Inertia multiplier & $\mathrm{LogU}[1,4]$ \\
            CoM offset & Up to $5$~mm per axis \\
            Friction multiplier & $\mathcal U[0.5,2.0]$ \\
            PD-gain multipliers & $\mathcal U[0.5,2.0]$ \\
            Wrist orientation & Random over the full orientation range \\
            External disturbances & $\sigma_a=6.67$~m/s$^2$; $\sigma_\alpha=133.33$~rad/s$^2$ \\
            Disturbance probability & $\mathrm{LogU}[1/3000,1/30]$ per step \\
            \midrule
            Joint-position noise & $\mathcal U[-0.02,0.02]$~rad \\
            Object-pose noise & $\sigma_p=10$~mm/axis; $\sigma_R=5^\circ$ (clip $15^\circ$) \\
            Pose delay & 0--3 steps (0--150~ms) \\
            Pose dropout & $3\%$ per step; hold last pose \\
            \bottomrule
        \end{tabular}
    \end{table}

    \section{Experiments}
    \label{sec:experiments}

    We use simulation and hardware experiments for complementary purposes.
    Simulation provides controlled, quantitative evidence for the three main
    design choices: diverse stable-grasp initialization, the adaptive goal
    curriculum, and the grasp-maintenance reward. Hardware experiments
    instead focus on system-level capabilities, including reaching across
    object geometries and wrist orientations, recovery from external
    disturbances, and continued operation after reaching.

    \subsection{Experimental Setup}
    \label{sec:experimental_setup}

    \noindent\textbf{Simulation.}
    We train the policies with asymmetric PPO in Isaac
    Lab~\cite{mittal2025isaac}. Physics is
    simulated at 120~Hz, and the policy produces incremental joint-position
    commands at 20~Hz. Both actor and critic are $[512,256,128]$ ELU MLPs, and
    object geometry is represented by a 64-dimensional BPS descriptor. Unless
    noted otherwise, ablations use a 40-mm Hexagonal Prism and fixed palm-up
    wrist, changing only the component under study. A target is considered
    reached when the position and orientation errors remain below 10~mm and
    $10^\circ$, respectively, for 20 consecutive policy steps. After reaching,
    a new target is issued without resetting the hand--object state. Each
    episode lasts 400 policy steps and terminates early if the object is
    dropped.

        \begin{figure}[t]
        \centering
        \includegraphics[width=0.8\columnwidth]{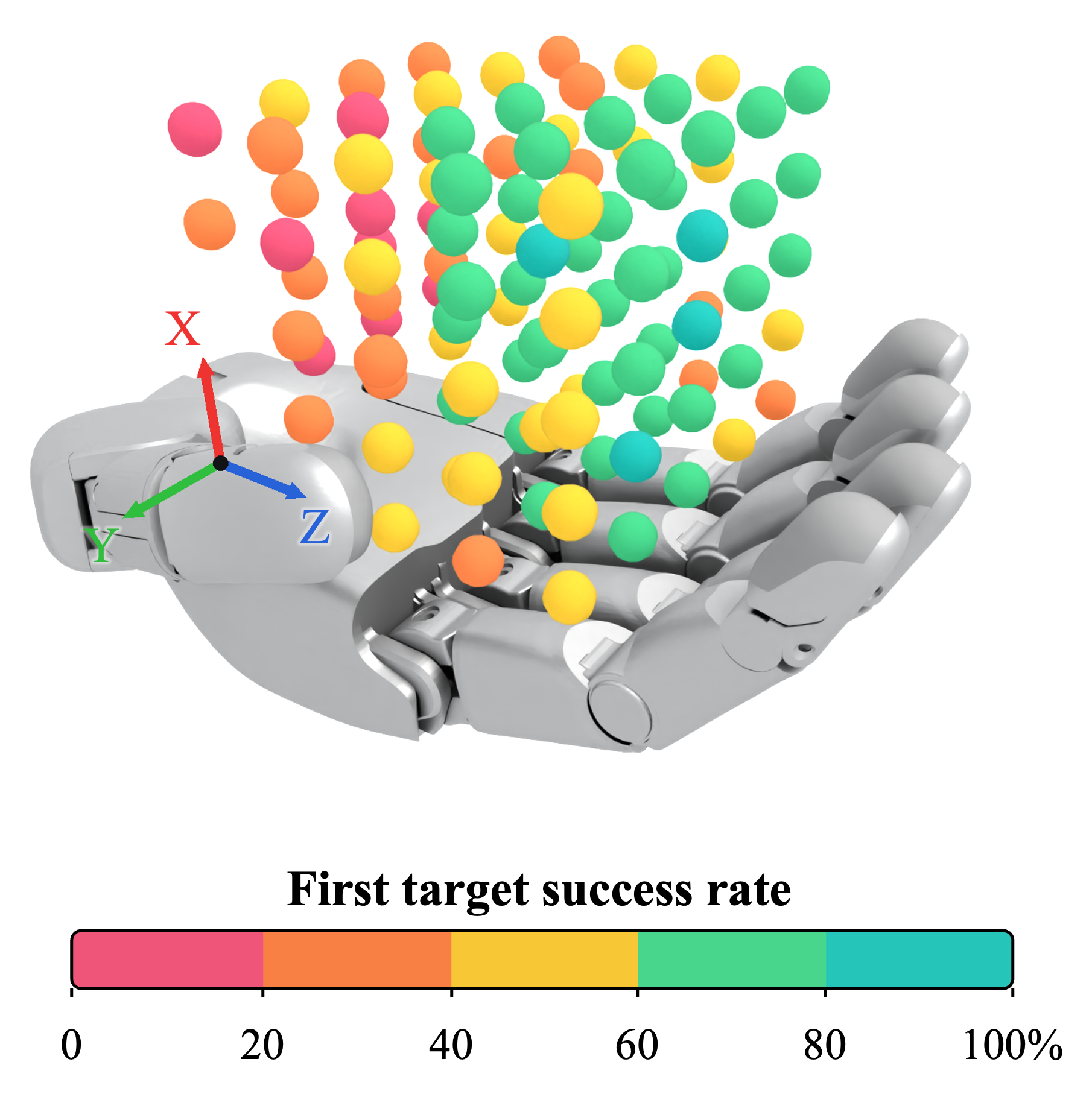}
        \caption{First-target success of the diverse policy across
        independently generated held-out initial grasps under paired
        $30$-mm/$180^\circ$ targets. Sphere positions indicate the initial
        object positions in the palm frame, and color indicates success rate.}
        \label{fig:grasp_heldout_workspace}
    \end{figure}

    \begin{figure}[t]
        \centering
        \includegraphics[width=\columnwidth]{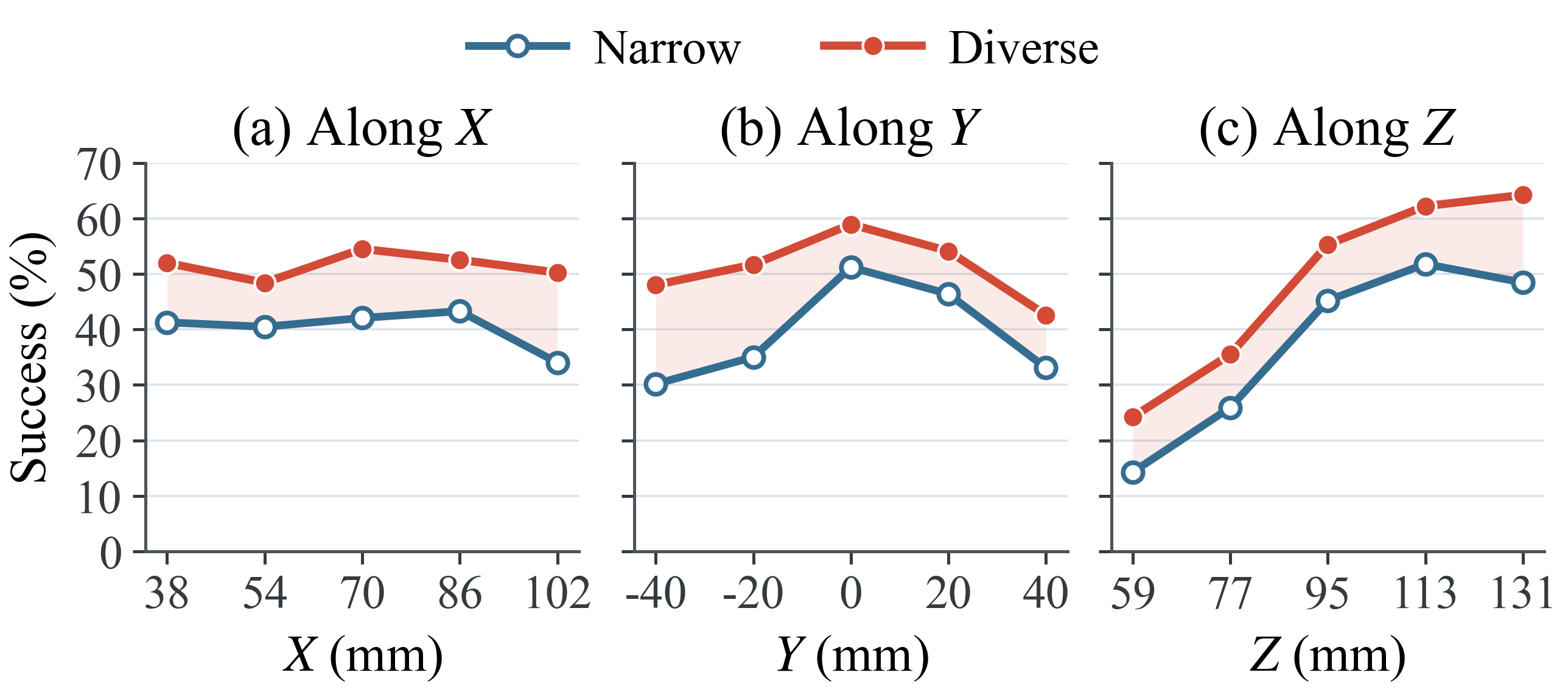}
        \caption{First-target success of the narrow and diverse policies,
        marginalized along each palm-frame position axis. Palm-frame $X$ is
        approximately normal to the palm, while $Y$ and $Z$ lie in the palm
        plane.}
        \label{fig:grasp_heldout_axis_trends}
    \end{figure}

    \begin{figure*}[t]
        \centering
        \includegraphics[width=\textwidth]{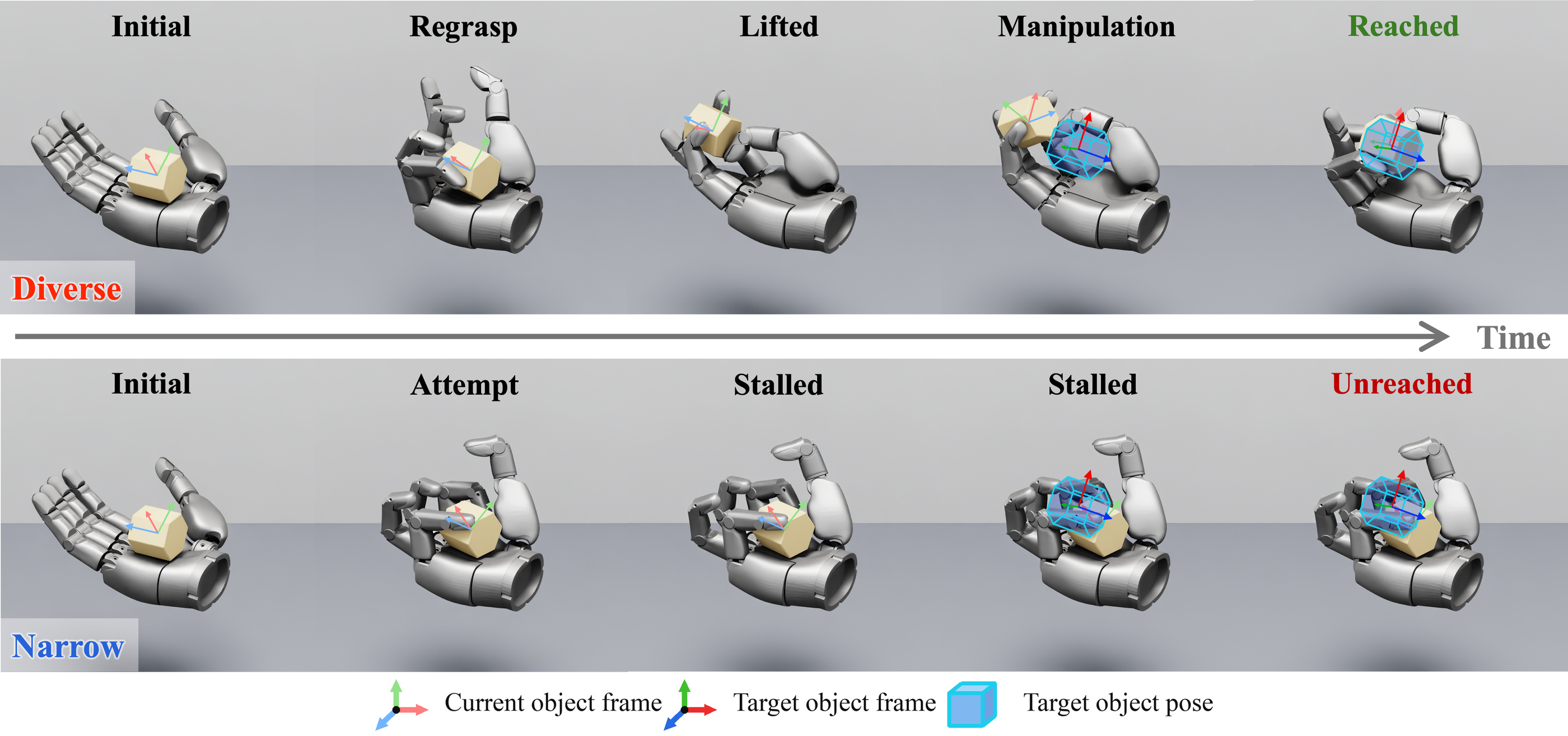}
        \caption{Simulation recovery from the same post-drop state and target.
        The diverse policy (top) rebuilds contact, lifts the object from the
        palm, and reaches the target. The narrow policy (bottom) attempts to
        regrasp the object but stalls before reaching the target.}
        \label{fig:sim_recovery}
    \end{figure*}

    \noindent\textbf{Hardware.}
    The real system consists of a 22-DoF Sharpa Wave hand~\cite{sharpaWave}
    and a RealSense D435 camera~\cite{realsenseD435}.
    FoundationPose~\cite{wen2024foundationpose} estimates the object pose from
    the RGB-D stream and a known object mesh. A calibrated camera-to-hand
    transform expresses the estimate
    in the palm frame, and the policy runs in closed loop at 20~Hz. The same
    observation and action interfaces are used in simulation and on hardware;
    only deployable proprioceptive and visual estimates are provided to the
    actor. The domain randomization used for transfer is summarized in
    Table~\ref{tab:sim2real_randomization}.

    \subsection{Quantitative Simulation Studies}
    \label{sec:simulation_experiments}

    \subsubsection{Generalization Across Initial Grasps}
    \label{sec:grasp_initialization_ablation}

    To isolate the effect of diversity in grasp initialization, we compare
    the proposed method with a narrow variant whose reset cache is obtained by
    perturbing the joint state and object pose of a single grasp seed. Each
    cache contains 6,842 stable reset states; the conditions differ only in
    their reset-state distributions. We test whether this coverage supports
    reaching from unseen stable grasps and recovery after the original grasp
    is lost.

    On independently generated held-out stable grasps, both policies receive
    the same initial states and $30$-mm/$180^\circ$ first targets. Diverse
    initialization increases first-target success from 40.1\% to 51.5\% and
    the mean number of completed goals per episode from 0.69 to 1.00.
    Figs.~\ref{fig:grasp_heldout_workspace} and
    \ref{fig:grasp_heldout_axis_trends} further characterize this gain over
    the initial object position. Smaller $Z$ coordinates and the boundaries
    of the $Y$ range are more difficult for both policies, while success varies
    less along $X$. The diverse policy has higher observed success in every
    marginalized position bin shown, suggesting that the gain is distributed
    across the held-out grasp workspace.

    The recovery test then examines whether this broader state coverage also
    improves recovery after a grasp is broken. Both policies are evaluated on the
    same 720 trials, with the object resting on the palm and no established
    grasp. For $30$-mm/$180^\circ$ targets that require lifting the object away
    from the palm, diverse initialization
    increases first-target success from 33.8\% to 72.9\%. Among successful
    trials, the mean time to recover and reach the target decreases from
    11.63~s to 9.45~s. Figure~\ref{fig:sim_recovery} contrasts representative
    rollouts from the same post-drop state and target: the diverse policy
    rebuilds multi-finger contact, lifts the object, and reaches the target,
    whereas the narrow policy stalls without reaching the target. Together, the
    two tests indicate that broader coverage during training improves both
    generalization across grasp poses and recovery after a drop.

    \subsubsection{Effect of the Adaptive Goal Curriculum}
    \label{sec:curriculum_ablation}

    We compare our curriculum with a variant that samples target magnitudes
    uniformly over the full range throughout training. Each training run uses
    64,000 parallel simulation environments. We evaluate the policies from
    the same stable grasps and with the same first targets in two suites:
    full-range targets with rotations in $[5^\circ,180^\circ]$ and translations
    in $[10,30]$~mm, and maximum-change targets of $180^\circ$ and $30$~mm.

    We report strict first-target success under the standard
    $10^\circ$/10-mm criterion and near success under a relaxed
    $20^\circ$/20-mm criterion. The goals-per-episode metric counts targets
    completed under the strict criterion before reset.

    \begin{table}[t]
        \centering
        \caption{Ablation of the adaptive goal curriculum on the hexagonal
        prism. Strict and near denote first-target success under the
        $10^\circ$/10-mm and $20^\circ$/20-mm criteria, respectively.}
        \label{tab:curriculum_ablation}
        \setlength{\tabcolsep}{2.0pt}
        \footnotesize
        \begin{tabular}{@{}lcccccc@{}}
            \toprule
            & \multicolumn{3}{c}{Full range}
            & \multicolumn{3}{c}{Maximum change} \\
            \cmidrule(lr){2-4}\cmidrule(l){5-7}
            Training & Strict & Near & Goals/ep. & Strict & Near & Goals/ep. \\
            \midrule
            Without curriculum
                & 6.2\% & 18.8\% & 0.08
                & 3.4\% & 13.9\% & 0.04 \\
            \textbf{Curriculum}
                & \textbf{59.5\%} & \textbf{77.0\%} & \textbf{1.55}
                & \textbf{55.3\%} & \textbf{72.8\%} & \textbf{1.08} \\
            \bottomrule
        \end{tabular}
    \end{table}

    As shown in Table~\ref{tab:curriculum_ablation}, directly training on the
    full goal range rarely produces successful 6D reaching. The curriculum
    raises strict success from 6.2\% to 59.5\% on full-range targets and from
    3.4\% to 55.3\% on maximum-change targets. Near success and the number of
    completed goals improve consistently as well. Thus, progressively expanding
    the goal range is important for learning large coupled translations and
    rotations.

    \subsubsection{Effect of the Grasp-Maintenance Reward}
    \label{sec:grasp_reward_ablation}

    We isolate the grasp-maintenance reward by setting only the weight of
    $r_t^{\mathrm{grasp}}$ to zero. We independently train each configuration
    three times. For each paired comparison, we average the metrics over
    equal-length training windows after both policies have reached the full
    $30$-mm/$180^\circ$ goal range, and report the mean and sample standard
    deviation across the three runs.

    We focus on four complementary metrics. Grasp-wrench quality $Q$ is the
    smooth minimum over the six bidirectional force and torque margins defined
    in Eq.~\eqref{eq:grasp_wrench_quality}. The minimum force and torque margins
    are the weakest bidirectional coverage values over the three force axes and
    three torque axes, respectively. Episode success rate is the fraction of
    completed training episodes in which at least one target is reached. Higher
    values are better for all four metrics.

    \begin{table}[t]
        \centering
        \caption{Grasp-maintenance reward ablation.}
        \label{tab:grasp_reward_ablation}
        \setlength{\tabcolsep}{3pt}
        \footnotesize
        \begin{tabular}{@{}lcc@{}}
            \toprule
            Metric & Without grasp reward & \textbf{With grasp reward} \\
            \midrule
            Grasp-wrench quality $Q$
                & $0.032\!\pm\!0.002$
                & $\mathbf{0.045\!\pm\!0.005}$ \\
            Minimum force margin
                & $0.077\!\pm\!0.003$
                & $\mathbf{0.107\!\pm\!0.010}$ \\
            Minimum torque margin
                & $0.036\!\pm\!0.002$
                & $\mathbf{0.047\!\pm\!0.004}$ \\
            Episode success rate (\%)
                & $53.7\!\pm\!2.5$
                & $\mathbf{59.1\!\pm\!2.5}$ \\
            \bottomrule
        \end{tabular}
    \end{table}

    As summarized in Table~\ref{tab:grasp_reward_ablation}, the grasp-maintenance reward increases
    quality by approximately 41\%, the minimum force margin by 39\%, and the
    minimum torque margin by 31\% on average. All three grasp metrics improve
    in every run. Episode success rate also rises from 53.7\% to 59.1\%, a mean
    gain of 5.4 percentage points. Thus, encouraging the policy to preserve
    balanced wrench resistance improves the resulting grasp without sacrificing
    pose-reaching performance.

    \subsection{Real-World Capability Evaluation}
    \label{sec:real_world_evaluation}

    We evaluate the complete visual-feedback system on hardware across object
    geometries and wrist orientations, under external disturbances, and during
    continued reaching without resets.

    \subsubsection{Reaching Across Object Geometries}
    \label{sec:real_objects}

    We evaluate the real-world system on four objects with distinct
    manipulation geometries: Cube, Hexagonal Prism, Square Bifrustum, and
    Hammer. The first three are drawn from the three training shape families
    and sizes and share one geometry-conditioned policy jointly trained on all
    nine shape--size combinations. They differ in symmetry and surface
    structure while having comparable scales. The Hammer, measuring
    $215\times50\times33.6$~mm, further tests the framework on an object with a
    substantially larger aspect ratio, moment arm, and translational workspace.
    Owing to its elongated geometry, the Hammer uses an extended translation
    curriculum of up to 100~mm, while the maximum rotation remains
    $180^\circ$.

\begin{figure}[t]
        \centering
        \includegraphics[width=\columnwidth]{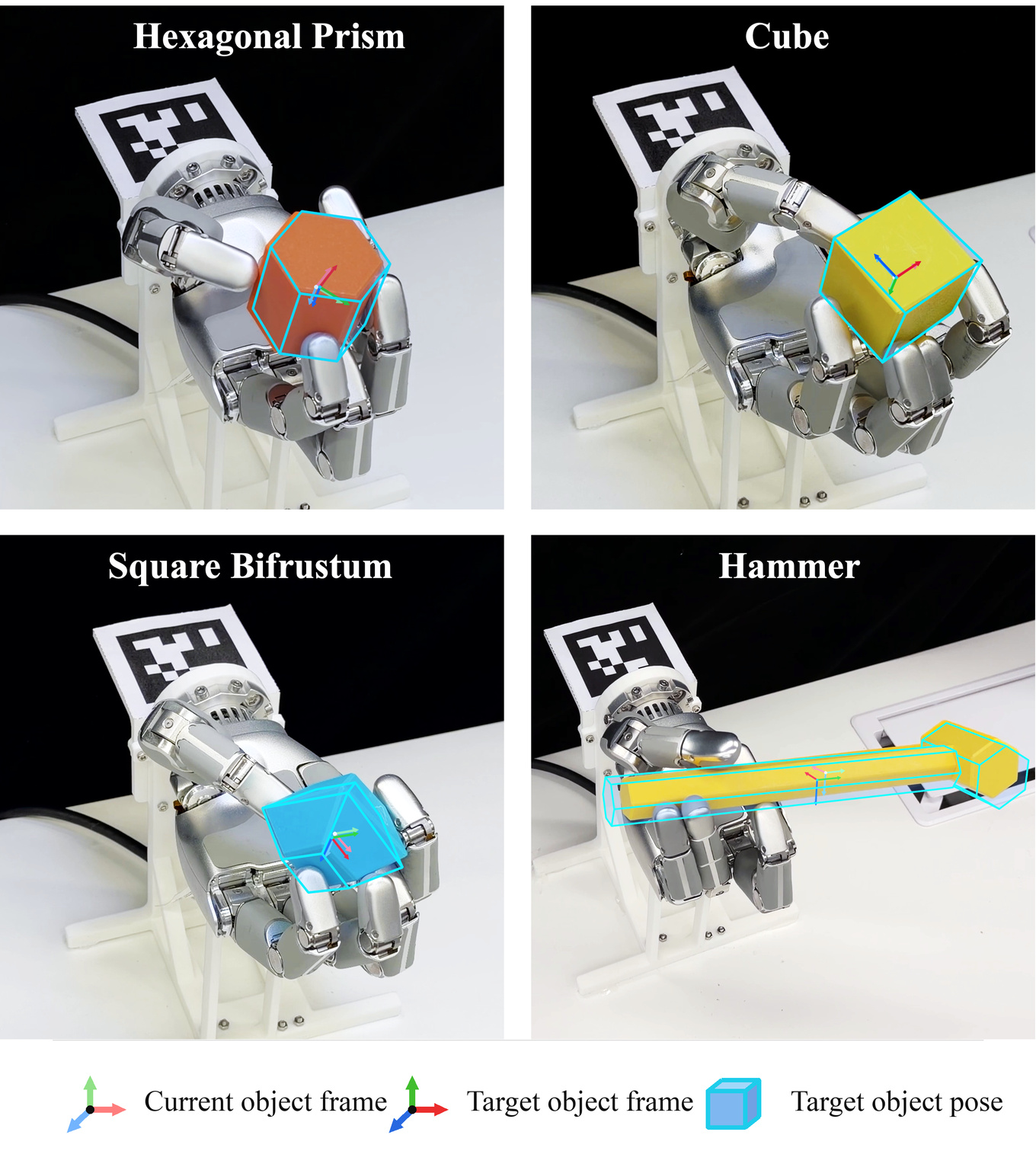}
        \caption{Representative snapshots after successful 6D pose reaching
        for the Hexagonal Prism, Cube, Square Bifrustum, and Hammer. The cyan
        outline denotes the commanded object pose, and the coordinate frames
        indicate the current and target poses.}
        \label{fig:real_objects}
    \end{figure}

    \noindent\textbf{Continuous random goal reaching.}
    For each object, we conduct uninterrupted rollouts in which 6D targets are
    sampled randomly from its feasible workspace. Once the position and
    orientation errors remain within their tolerances for the required dwell
    time, a new target is issued without resetting the hand--object system.
    In each rollout, POISE completed at least
    five successive goals without resetting the hand--object system.
    The experiment therefore evaluates not only whether the policy can reach
    an individual pose, but also whether it retains sufficient control of the
    object for subsequent reaching. Figure~\ref{fig:real_objects} shows one
    representative reached pose for each object geometry.

    \noindent\textbf{Reaching user-specified targets.}
    We further test whether the system can follow explicitly specified goals
    rather than only targets drawn from the random sampler. For the Hexagonal
    Prism, the commanded sequence moves the object to different in-hand
    positions while successively orienting different prism faces upward. For
    the Hammer, we prescribe full 6D targets involving substantial translation
    and changes in its principal-axis direction. In the representative
    sequences, POISE reaches four Hexagonal Prism targets with changes of up to
    35~mm and $100^\circ$, and seven Hammer targets with changes of up to
    93~mm and $180^\circ$. The corresponding mean times to reach a target
    are 3.0~s and 2.7~s. In both cases, each new target is issued from the
    attained state without resetting the hand or object. Representative
    reaching sequences are shown in Fig.~\ref{fig:real_target_tracking}.

    \subsubsection{Reaching Under Different Wrist Orientations}
    \label{sec:real_wrist_orientations}

\begin{figure}[t]
        \centering
        \includegraphics[width=\columnwidth]{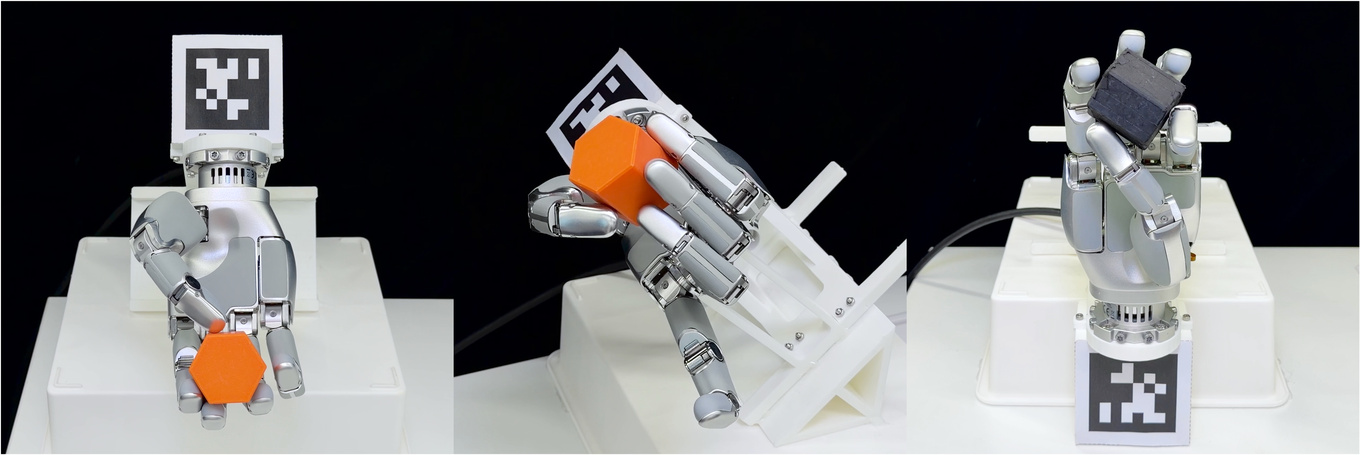}
        \caption{Representative real-world reaches under three wrist
        orientations with different gravity directions in the palm frame.}
        \label{fig:real_wrist}
    \end{figure}

    We deploy the same policy with the wrist fixed in three orientations,
    thereby changing gravity in the palm frame and reducing the passive
    support available from the palm. Nevertheless, the policy sustains
    continued reaching in every configuration. In the representative rollouts
    shown from left to right in Fig.~\ref{fig:real_wrist}, it completes 7, 5,
    and 6 successive goals without reset, with mean reaching times of 5.1,
    5.5, and 7.1~s. Across the three conditions, the per-target minimum error
    averages 5.3--7.4~mm in position and $3.2^\circ$--$4.4^\circ$ in
    orientation, based on the online pose estimates. These results demonstrate
    continued reaching under different gravity directions using one policy.

    \subsubsection{Closed-Loop Recovery from Disturbances}
    \label{sec:real_disturbance_recovery}

\begin{figure*}[t]
        \centering
        \includegraphics[width=\textwidth]{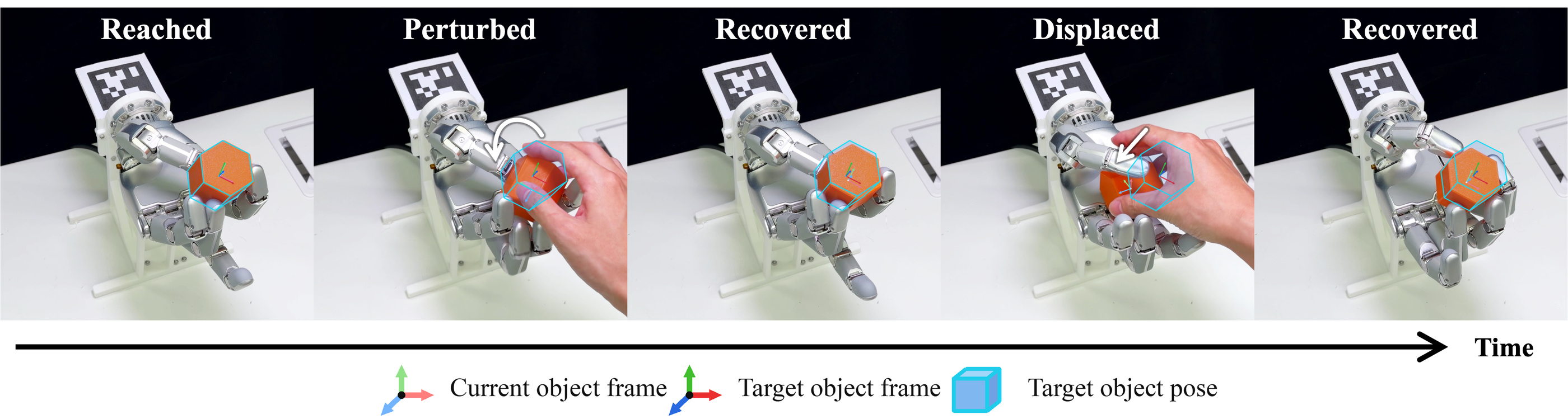}
        \caption{Closed-loop recovery from two consecutive disturbances on
        hardware. After reaching the target, the object is first perturbed in
        orientation and then displaced more substantially, changing its
        contacts. The policy
        returns it to the unchanged target after each disturbance.}
        \label{fig:real_recovery}
    \end{figure*}

    We further evaluate whether the policy can recover from unexpected
    disturbances during real-world operation. After the object reaches the
    commanded pose, an experimenter perturbs it twice while keeping the target
    unchanged. As shown in Fig.~\ref{fig:real_recovery}, the
    first perturbation primarily changes the object orientation, whereas the
    second produces a larger displacement and a change in the contact
    configuration. In both cases,
    the policy responds to the updated visual pose estimate, reorganizes the
    contacts, and returns the object to the commanded pose. This sequence
    demonstrates closed-loop recovery from externally induced pose and contact
    deviations.

    \subsubsection{Hardware Ablation of the Grasp-Maintenance Reward}
    \label{sec:real_grasp_reward_ablation}

    We deploy the two policies from Sec.~\ref{sec:grasp_reward_ablation} in
    10 trials each. Starting from the same grasp, each policy attempts the
    same sequence of three targets without reset; completing all three defines
    sequence success. Reach time and steady-state errors are reported as
    medians over reached targets, with errors averaged over the final 20 policy
    steps.

    \begin{table}[t]
        \centering
        \caption{Real-world reaching success with and without the grasp reward.}
        \label{tab:hardware_grasp_ablation}
        \setlength{\tabcolsep}{3.5pt}
        \footnotesize
        \begin{tabular}{@{}lcc@{}}
            \toprule
            Metric & Without grasp reward & \textbf{With grasp reward} \\
            \midrule
            Complete sequences & $2/10$ & $\mathbf{8/10}$ \\
            Planned targets completed & $8/30$ & $\mathbf{27/30}$ \\
            Mean targets reached & $0.8$ & $\mathbf{2.7}$ \\
            Reach time (s) & $6.8$ & $\mathbf{4.9}$ \\
            Position error (mm) & $4.8$ & $\mathbf{6.4}$ \\
            Orientation error ($^\circ$) & $4.7$ & $\mathbf{5.9}$ \\
            \bottomrule
        \end{tabular}
    \end{table}

    As shown in Table~\ref{tab:hardware_grasp_ablation}, the reward raises
    sequence success from $2/10$ to $8/10$ and target success from 26.7\% to
    90.0\%, while maintaining comparable pose accuracy. Without the reward,
    progressive contact loss often led to object drops. The full policy instead
    tended to retain more distributed multi-finger contacts, improving the
    reliability of continued reaching.

    \section{Conclusion and Future Work}
    \label{sec:conclusion}
    \label{sec:limitations}

    We presented POISE, a sim-to-real RL framework for reaching palm-relative
    6D object poses through finger motions alone. Our results show that diverse
    stable-grasp initialization improves generalization to unseen grasp
    configurations and recovery after contact loss, while the adaptive goal
    curriculum and grasp-preserving reward enable stable, large-range pose
    reaching. Together, these findings establish palm-relative
    $\mathrm{SE}(3)$ reaching as a useful primitive for functional in-hand
    reconfiguration, with the potential to support tool use and other
    downstream tasks as a step toward general-purpose dexterous manipulation.

    \noindent\textbf{Limitations and future work.}
    Two limitations remain. First, POISE specifies the object target pose but
    not a desired hand or contact configuration. The policy may therefore use
    effective but unnatural hand postures. Conditioning on both object and
    grasp targets could produce more natural, task-appropriate motions. Second,
    real-world performance still lags that in simulation because of imperfect contact
    modeling and errors in 6D pose tracking. Future work will reduce this gap by
    incorporating online point-cloud observations through observation
    distillation, adding tactile feedback, and adapting from real interactions,
    for example by learning residual dynamics.

\bibliography{reference}

@inproceedings{dafle2018motioncones,
  author    = {Dafle, Nikhil Chavan and Holladay, Rachel and Rodriguez, Alberto},
  title     = {In-Hand Manipulation via Motion Cones},
  booktitle = {Proceedings of Robotics: Science and Systems},
  year      = {2018},
  address   = {Pittsburgh, Pennsylvania},
  month     = {June},
  doi       = {10.15607/RSS.2018.XIV.058}
}

@inproceedings{han1998rolling,
  author    = {Han, Li and Trinkle, Jeffrey C.},
  title     = {Dextrous Manipulation by Rolling and Finger Gaiting},
  booktitle = {Proceedings of the 1998 IEEE International Conference on Robotics and Automation},
  volume    = {1},
  pages     = {730--735},
  year      = {1998},
  doi       = {10.1109/ROBOT.1998.677060}
}

@article{sundaralingam2019relaxed,
  author  = {Sundaralingam, Balakumar and Hermans, Tucker},
  title   = {Relaxed-Rigidity Constraints: Kinematic Trajectory Optimization and Collision Avoidance for In-Grasp Manipulation},
  journal = {Autonomous Robots},
  volume  = {43},
  number  = {2},
  pages   = {469--483},
  year    = {2019},
  doi     = {10.1007/s10514-018-9772-z}
}

@article{morgan2022complex,
  author  = {Morgan, Andrew S. and Hang, Kaiyu and Wen, Bowen and Bekris, Kostas and Dollar, Aaron M.},
  title   = {Complex In-Hand Manipulation Via Compliance-Enabled Finger Gaiting and Multi-Modal Planning},
  journal = {IEEE Robotics and Automation Letters},
  volume  = {7},
  number  = {2},
  pages   = {4821--4828},
  year    = {2022},
  doi     = {10.1109/LRA.2022.3145961}
}

@inproceedings{chanrungmaneekul2023nonparametric,
  author    = {Chanrungmaneekul, Podshara and Ren, Kejia and Grace, Joshua T. and Dollar, Aaron M. and Hang, Kaiyu},
  title     = {Non-Parametric Self-Identification and Model Predictive Control of Dexterous In-Hand Manipulation},
  booktitle = {2023 IEEE/RSJ International Conference on Intelligent Robots and Systems (IROS)},
  pages     = {8743--8750},
  year      = {2023},
  organization = {IEEE}
}

@inproceedings{jiang2024contact,
  author    = {Jiang, Yongpeng and Yu, Mingrui and Zhu, Xinghao and Tomizuka, Masayoshi and Li, Xiang},
  title     = {Contact-Implicit Model Predictive Control for Dexterous In-Hand Manipulation: A Long-Horizon and Robust Approach},
  booktitle = {2024 IEEE/RSJ International Conference on Intelligent Robots and Systems (IROS)},
  pages     = {5260--5266},
  year      = {2024},
  publisher = {IEEE},
  doi       = {10.1109/IROS58592.2024.10801751}
}

@article{andrychowicz2020learning,
  author  = {Andrychowicz, Marcin and Baker, Bowen and Chociej, Maciek and Jozefowicz, Rafal and McGrew, Bob and Pachocki, Jakub and Petron, Arthur and Plappert, Matthias and Powell, Glenn and Ray, Alex and Schneider, Jonas and Sidor, Szymon and Tobin, Josh and Welinder, Peter and Weng, Lilian and Zaremba, Wojciech},
  title   = {Learning Dexterous In-Hand Manipulation},
  journal = {The International Journal of Robotics Research},
  volume  = {39},
  number  = {1},
  pages   = {3--20},
  year    = {2020},
  doi     = {10.1177/0278364919887447}
}

@inproceedings{chen2022system,
  author    = {Chen, Tao and Xu, Jie and Agrawal, Pulkit},
  title     = {A System for General In-Hand Object Re-Orientation},
  booktitle = {Proceedings of the 5th Conference on Robot Learning},
  pages     = {297--307},
  year      = {2022},
  volume    = {164},
  series    = {Proceedings of Machine Learning Research},
  publisher = {PMLR}
}

@inproceedings{qi2023hora,
  author    = {Qi, Haozhi and Kumar, Ashish and Calandra, Roberto and Ma, Yi and Malik, Jitendra},
  title     = {In-Hand Object Rotation via Rapid Motor Adaptation},
  booktitle = {Proceedings of the 6th Conference on Robot Learning},
  pages     = {1722--1732},
  year      = {2023},
  volume    = {205},
  series    = {Proceedings of Machine Learning Research},
  publisher = {PMLR}
}

@inproceedings{qi2023rotateit,
  author    = {Qi, Haozhi and Yi, Brent and Suresh, Sudharshan and Lambeta, Mike and Ma, Yi and Calandra, Roberto and Malik, Jitendra},
  title     = {General In-Hand Object Rotation with Vision and Touch},
  booktitle = {Proceedings of the 7th Conference on Robot Learning},
  pages     = {2549--2564},
  year      = {2023},
  volume    = {229},
  series    = {Proceedings of Machine Learning Research},
  publisher = {PMLR}
}

@article{akkaya2019rubiks,
  author  = {Akkaya, Ilge and Andrychowicz, Marcin and Chociej, Maciek and Litwin, Mateusz and McGrew, Bob and Petron, Arthur and Paino, Alex and Plappert, Matthias and Powell, Glenn and Ribas, Raphael and others},
  title   = {Solving {Rubik's} Cube with a Robot Hand},
  journal = {arXiv preprint arXiv:1910.07113},
  year    = {2019}
}

@misc{huang2021geometry,
  author        = {Huang, Wenlong and Mordatch, Igor and Abbeel, Pieter and Pathak, Deepak},
  title         = {Generalization in Dexterous Manipulation via Geometry-Aware Multi-Task Learning},
  year          = {2021},
  eprint        = {2111.03062},
  howpublished  = {arXiv preprint arXiv:2111.03062},
  archivePrefix = {arXiv},
  primaryClass  = {cs.RO}
}

@article{plappert2018multigoal,
  author  = {Plappert, Matthias and Andrychowicz, Marcin and Ray, Alex and McGrew, Bob and Baker, Bowen and Powell, Glenn and Schneider, Jonas and Tobin, Josh and Chociej, Maciek and Welinder, Peter and Kumar, Vikash and Zaremba, Wojciech},
  title   = {Multi-Goal Reinforcement Learning: Challenging Robotics Environments and Request for Research},
  journal = {arXiv preprint arXiv:1802.09464},
  year    = {2018}
}

@inproceedings{charlesworth2021solving,
  author    = {Charlesworth, Henry J. and Montana, Giovanni},
  title     = {Solving Challenging Dexterous Manipulation Tasks With Trajectory Optimisation and Reinforcement Learning},
  booktitle = {Proceedings of the 38th International Conference on Machine Learning},
  pages     = {1496--1506},
  year      = {2021},
  volume    = {139},
  series    = {Proceedings of Machine Learning Research},
  publisher = {PMLR}
}

@misc{kedia2026simtoolreal,
  author        = {Kedia, Kushal and Lum, Tyler Ga Wei and Bohg, Jeannette and Liu, C. Karen},
  title         = {{SimToolReal}: An Object-Centric Policy for Zero-Shot Dexterous Tool Manipulation},
  year          = {2026},
  eprint        = {2602.16863},
  howpublished  = {arXiv preprint arXiv:2602.16863},
  archivePrefix = {arXiv},
  primaryClass  = {cs.RO}
}

@misc{lum2026play2perfect,
  author        = {Lum, Tyler Ga Wei and Kedia, Kushal and Liu, C. Karen and Bohg, Jeannette},
  title         = {{Play2Perfect}: What Matters in Dexterous Play Pretraining for Precise Assembly?},
  year          = {2026},
  eprint        = {2606.26428},
  howpublished  = {arXiv preprint arXiv:2606.26428},
  archivePrefix = {arXiv},
  primaryClass  = {cs.RO}
}

@article{yin2025lightning,
  author  = {Yin, Zhao-Heng and Abbeel, Pieter},
  title   = {{Lightning Grasp}: High Performance Procedural Grasp Synthesis with Contact Fields},
  journal = {arXiv preprint arXiv:2511.07418},
  year    = {2025}
}

@inproceedings{prokudin2019efficient,
  author    = {Prokudin, Sergey and Lassner, Christoph and Romero, Javier},
  title     = {Efficient Learning on Point Clouds With Basis Point Sets},
  booktitle = {2019 IEEE/CVF International Conference on Computer Vision (ICCV)},
  pages     = {4331--4340},
  year      = {2019},
  publisher = {IEEE},
  doi       = {10.1109/ICCV.2019.00443}
}

@inproceedings{pitz2024shapeconditioned,
  author    = {Pitz, Johannes and R{\"o}stel, Lennart and Sievers, Leon and Burschka, Darius and B{\"a}uml, Berthold},
  title     = {Learning a Shape-Conditioned Agent for Purely Tactile In-Hand Manipulation of Various Objects},
  booktitle = {2024 IEEE/RSJ International Conference on Intelligent Robots and Systems (IROS)},
  pages     = {13112--13119},
  year      = {2024},
  publisher = {IEEE},
  doi       = {10.1109/IROS58592.2024.10802864}
}

@INPROCEEDINGS{qi2025translation,
  author={Yin, Jessica and Qi, Haozhi and Malik, Jitendra and Pikul, James and Yim, Mark and Hellebrekers, Tess},
  booktitle={2025 IEEE International Conference on Robotics and Automation (ICRA)},
  title={Learning In-Hand Translation Using Tactile Skin with Shear and Normal Force Sensing},
  year={2025},
  volume={},
  number={},
  pages={5850--5856},
  doi={10.1109/ICRA55743.2025.11127974}}

@inproceedings{liu2026dexndm,
 author = {Liu, Xueyi and Wang, He and Yi, Li},
 booktitle = {International Conference on Learning Representations},
 title = {{DexNDM}: Closing the Reality Gap for Dexterous In-Hand Rotation via Joint-Wise Neural Dynamics Model},
 year = {2026},
 url = {https://openreview.net/forum?id=80vjyj5o7l}
}

@misc{schulman2017ppo,
      title={Proximal Policy Optimization Algorithms},
      author={John Schulman and Filip Wolski and Prafulla Dhariwal and Alec Radford and Oleg Klimov},
      year={2017},
      eprint={1707.06347},
      archivePrefix={arXiv},
      primaryClass={cs.LG},
      url={https://arxiv.org/abs/1707.06347},
}

@INPROCEEDINGS{wen2024foundationpose,
  author={Wen, Bowen and Yang, Wei and Kautz, Jan and Birchfield, Stan},
  booktitle={2024 IEEE/CVF Conference on Computer Vision and Pattern Recognition (CVPR)},
  title={{FoundationPose}: Unified {6D} Pose Estimation and Tracking of Novel Objects},
  year={2024},
  volume={},
  number={},
  pages={17868--17879},
  doi={10.1109/CVPR52733.2024.01692}}

@misc{sharpaWave,
  author       = {{Sharpa}},
  title        = {{Sharpa Wave}},
  url          = {https://www.sharpa.com/pages/wave},
  note         = {Accessed: Aug. 27, 2026}
}

@misc{realsenseD435,
  author       = {{RealSense}},
  title        = {{RealSense D435}},
  url          = {https://www.realsenseai.com/cn/products/stereo-depth-camera-d435/},
  note         = {Accessed: Aug. 27, 2026}
}

@article{mittal2025isaac,
  title={{Isaac Lab}: A {GPU}-accelerated simulation framework for multi-modal robot learning},
  author={Mittal, Mayank and Roth, Pascal and Tigue, James and Richard, Antoine and Zhang, Octi and Du, Peter and Serrano-Munoz, Antonio and Yao, Xinjie and Zurbr{\"u}gg, Ren{\'e} and Rudin, Nikita and others},
  journal={arXiv preprint arXiv:2511.04831},
  year={2025}
}

@misc{yang2024anyrotate,
      title={{AnyRotate}: Gravity-Invariant In-Hand Object Rotation with Sim-to-Real Touch},
      author={Max Yang and Chenghua Lu and Alex Church and Yijiong Lin and Chris Ford and Haoran Li and Efi Psomopoulou and David A. W. Barton and Nathan F. Lepora},
      year={2024},
      eprint={2405.07391},
      archivePrefix={arXiv},
      primaryClass={cs.RO},
      url={https://arxiv.org/abs/2405.07391},
}

@misc{yin2025dexteritygen,
      title={{DexterityGen}: Foundation Controller for Unprecedented Dexterity},
      author={Zhao-Heng Yin and Changhao Wang and Luis Pineda and Francois Hogan and Krishna Bodduluri and Akash Sharma and Patrick Lancaster and Ishita Prasad and Mrinal Kalakrishnan and Jitendra Malik and Mike Lambeta and Tingfan Wu and Pieter Abbeel and Mustafa Mukadam},
      year={2025},
      eprint={2502.04307},
      archivePrefix={arXiv},
      primaryClass={cs.RO},
      url={https://arxiv.org/abs/2502.04307},
}

@article{zhang2026unicross,
  author  = {Zhang, Hui and Ferchow, Julian and Song, Jie and Meboldt, Mirko},
  title   = {{UniCross}: Unified Cross-Skill Dexterous Manipulation Synthesis},
  journal = {arXiv preprint arXiv:2607.28198},
  year    = {2026}
}

@inproceedings{liu2025dextrack,
  author    = {Liu, Xueyi and Adalibieke, Jianibieke and Han, Qianwei and Qin, Yuzhe and Yi, Li},
  title     = {{DexTrack}: Towards Generalizable Neural Tracking Control for Dexterous Manipulation from Human References},
  booktitle = {International Conference on Learning Representations},
  year      = {2025}
}

@inproceedings{li2025maniptrans,
  author    = {Li, Kailin and Li, Puhao and Liu, Tengyu and Li, Yuyang and Huang, Siyuan},
  title     = {{ManipTrans}: Efficient Dexterous Bimanual Manipulation Transfer via Residual Learning},
  booktitle = {Proceedings of the IEEE/CVF Conference on Computer Vision and Pattern Recognition (CVPR)},
  pages     = {6991--7003},
  year      = {2025}
}

@article{li2026teledexter,
  author  = {Li, Puhao and Chen, Zeyuan and Wu, Yingying and Wei, Pengkun and Li, Yuyang and Wang, Tianyu and Shi, Jiaxiao and Yu, Mingrui and Jia, Baoxiong and Zhu, Song-Chun and Liu, Tengyu and Huang, Siyuan},
  title   = {Towards Human-level Dexterous Teleoperation},
  journal = {arXiv preprint arXiv:2607.11481},
  year    = {2026}
}

@article{wang2025handobjecttracking,
  author  = {Wang, Yinhuai and Yu, Runyi and Tsui, Hok Wai and Lin, Xiaoyi and Zhang, Hui and Zhao, Qihan and Fan, Ke and Li, Miao and Song, Jie and Wang, Jingbo and Chen, Qifeng and Tan, Ping},
  title   = {Learning Generalizable Hand-Object Tracking from Synthetic Demonstrations},
  journal = {arXiv preprint arXiv:2512.19583},
  year    = {2025}
}

@inproceedings{yin2023rotating,
  author    = {Yin, Zhao-Heng and Huang, Binghao and Qin, Yuzhe and Chen, Qifeng and Wang, Xiaolong},
  title     = {Rotating without Seeing: Towards In-hand Dexterity through Touch},
  booktitle = {Proceedings of Robotics: Science and Systems},
  year      = {2023},
  doi       = {10.15607/RSS.2023.XIX.036}
}

@article{chen2023visual,
  title={Visual dexterity: In-hand reorientation of novel and complex object shapes},
  author={Chen, Tao and Tippur, Megha and Wu, Siyang and Kumar, Vikash and Adelson, Edward and Agrawal, Pulkit},
  journal={Science Robotics},
  volume={8},
  number={84},
  pages={eadc9244},
  year={2023},
  publisher={American Association for the Advancement of Science}
}

@inproceedings{handa2023dextreme,
  title={{DeXtreme}: Transfer of agile in-hand manipulation from simulation to reality},
  author={Handa, Ankur and Allshire, Arthur and Makoviychuk, Viktor and Petrenko, Aleksei and Singh, Ritvik and Liu, Jingzhou and Makoviichuk, Denys and Van Wyk, Karl and Zhurkevich, Alexander and Sundaralingam, Balakumar and others},
  booktitle={2023 IEEE International Conference on Robotics and Automation (ICRA)},
  pages={5977--5984},
  year={2023},
  organization={IEEE}
}

@inproceedings{khandate2023sampling,
  author    = {Khandate, Gagan and Shang, Siqi and Chang, Eric T. and Saidi, Tristan L. and Adams, Johnson and Ciocarlie, Matei},
  title     = {Sampling-based Exploration for Reinforcement Learning of Dexterous Manipulation},
  booktitle = {Proceedings of Robotics: Science and Systems},
  year      = {2023},
  doi       = {10.15607/RSS.2023.XIX.020}
}

@article{suh2026contacttrust,
  author  = {Suh, H. J. Terry and Pang, Tao and Zhao, Tong and Tedrake, Russ},
  title   = {Dexterous Contact-Rich Manipulation via the Contact Trust Region},
  journal = {The International Journal of Robotics Research},
  volume  = {45},
  number  = {9},
  pages   = {1418--1454},
  year    = {2026},
  doi     = {10.1177/02783649251398875}
}

@inproceedings{wang2023dexgraspnet,
  author    = {Wang, Ruicheng and Zhang, Jialiang and Chen, Jiayi and Xu, Yinzhen and Li, Puhao and Liu, Tengyu and Wang, He},
  title     = {{DexGraspNet}: A Large-Scale Robotic Dexterous Grasp Dataset for General Objects Based on Simulation},
  booktitle = {2023 IEEE International Conference on Robotics and Automation (ICRA)},
  pages     = {11359--11366},
  year      = {2023},
  doi       = {10.1109/ICRA48891.2023.10160982}
}

@misc{kuang2026dex4d,
      title={{Dex4D}: Task-Agnostic Point Track Policy for Sim-to-Real Dexterous Manipulation},
      author={Yuxuan Kuang and Sungjae Park and Katerina Fragkiadaki and Shubham Tulsiani},
      year={2026},
      eprint={2602.15828},
      archivePrefix={arXiv},
      primaryClass={cs.RO},
      url={https://arxiv.org/abs/2602.15828},
}

@inproceedings{bhardwaj2026viserdex,
  author    = {Bhardwaj, Arjun and Wilder-Smith, Maximum and Mittal, Mayank and Patil, Vaishakh and Hutter, Marco},
  title     = {{ViserDex}: Visual Sim-to-Real for Robust Dexterous In-hand Reorientation},
  booktitle = {Proceedings of Robotics: Science and Systems},
  year      = {2026}
}

@inproceedings{yin2026wmcraftnet,
  author    = {Yin, Jie and Zhao, Zeyuan and Tan, Xiaojing and Liu, Yang and Wang, Chiyu and Gu, Xinyang},
  title     = {{WM-Craftnet}: World Synesthesia Model for Generalizable and Robust Dexterous In-Hand Manipulation},
  booktitle = {Conference on Robot Learning},
  year      = {2026}
}

@inproceedings{pei2026assembling,
  author    = {Pei, Liuao and Wu, Tianyue and Zhang, Hui and Luo, Ping and Song, Jie},
  title     = {Assembling Two Parts in One Hand},
  booktitle = {Conference on Robot Learning},
  year      = {2026},
  note      = {to appear}
}
\end{document}